\documentclass[runningheads]{llncs}

\usepackage{tikz}
\usepackage{graphicx}

\usepackage{xcolor}
\PassOptionsToPackage{hyphens,spaces}{url}
\usepackage[colorlinks,urlcolor=blue]{hyperref}
\usepackage{xspace}
\usepackage{bm}
\usepackage[utf8]{inputenc}
\usepackage{amssymb}
\usepackage{amsfonts}
\usepackage{amsmath}

\usepackage{graphicx}
\usepackage{booktabs}
\usepackage{footnoterange}
\usepackage{multirow}
\usepackage{bbding}

\usepackage[ruled,linesnumbered,vlined]{algorithm2e}

\usepackage{numprint}
\npthousandsep{,}
\npdecimalsign{.}

\usepackage[binary-units=true,detect-weight=true]{siunitx}

\providecommand{\qty}{\SI}

  \newcommand*{\grk}[1]{\ensuremath{\mathsf{grk}_{#1}}}
\newcommand*{\casc}[1]{\ensuremath{\mathsf{casc}_{#1}}}

\newcommand*{\minFof}{$\mathsf{min}_\mathit{fof}$\xspace}
\newcommand*{\minCnfA}{$\mathsf{min}_\mathit{cnf1}$\xspace}
\newcommand*{\minCnfB}{$\mathsf{min}_\mathit{cnf2}$\xspace}
\newcommand*{\sMirek}{\textsf{gnn}\xspace}
\newcommand*{\sPepa}{\textsf{lgbm}\xspace}
\newcommand*{\sBushy}{\textsf{bushy}\xspace}

\begin{document}
\title{Solving Hard Mizar Problems with\\ Instantiation and Strategy Invention
}
\titlerunning{Solving Hard Mizar Problems}
\author{Jan Jakub\r{u}v\inst{1,2}\Envelope\orcidID{0000-0002-8848-5537} 
   \and  Mikol\'a\v{s} Janota\inst{1}\orcidID{0000-0003-3487-784X}
   \and Josef Urban\inst{1}\orcidID{0000-0002-1384-1613}
}
\authorrunning{J. Jakub\r{u}v et al.}
\institute{
  Czech Technical University in Prague, Czechia \\
  \email{Mikolas.Janota@cvut.cz}\quad
  \email{josef.urban@gmail.com}
\and
University of Innsbruck, Austria \\
  \email{jakubuv@gmail.com} 
}

\maketitle              %
\begin{abstract}
   In this work, we prove over 3000 previously ATP-unproved Mizar/MPTP problems
   by using several ATP and AI methods, raising the number of ATP-solved Mizar
   problems from 75\% to above 80\%.
  First, we start to
  experiment with the cvc5 SMT solver which uses several
  instantiation-based heuristics that differ from the
  superposition-based systems, 
  that were previously applied to Mizar,
  and add many new solutions. Then we use
  automated strategy invention to develop cvc5 strategies that largely
  improve cvc5's performance on the hard problems. In particular, the
  best invented strategy solves over 14\% more problems than the best
  previously available cvc5 strategy. We also show that different
  clausification methods have a high impact on such
  instantiation-based methods, again producing many new solutions. In
  total, 
  the methods solve \numprint{3021} (21.3\%) 
  of the \numprint{14163} previously unsolved hard Mizar problems.
  This is a new milestone over the Mizar large-theory
  benchmark and a large strengthening of the hammer methods
  for Mizar.

\end{abstract}

\section{Introduction: Mizar, ATPs, Hammers}

The Mizar Mathematical Library (MML)~\cite{BancerekBGKMNP18} is one of the earliest large libraries of formal mathematics, containing a wide selection of lemmas and theorems from various areas of mathematics.
The MML and the Mizar
system~\cite{mizar-first-30,BancerekBGKMNPU15,mizar-in-a-nutshell} has
been used as a source of automated theorem proving
(ATP)~\cite{DBLP:books/el/RobinsonV01} problems for over 25 years,
starting with the export of several Mizar articles done by the ILF
system~\cite{Dah97,Dahn98}. Since 2003, the MPTP
system~\cite{Urb04-MPTP0,Urban06} has been used to export the MML in the
DFG~\cite{Hahnle96} and later TPTP~\cite{SutcliffeSY94} formats.  In the
earliest (2003) ATP experiments over the whole library, 
state-of-the-art ATPs could prove about 40\% of these problems when their premises were limited to
those used in the human-written Mizar proofs (the so called \emph{bushy}\footnote{\url{https://tptp.org/MPTPChallenge}}, i.e., easier, mode).

Since 2013, a fixed version of the MML (1147) and MPTP  consisting of 57880
problems has been used as a large benchmark for ATPs and related
hammer~\cite{hammers4qed} (large-theory) methods over
Mizar~\cite{DBLP:conf/frocos/RawsonR19,JakubuvU19,DBLP:conf/frocos/Suda21,DBLP:conf/tableaux/RawsonR21,DBLP:journals/corr/abs-2303-15642,ChvalovskyKPU23}. When
using many ATP and premise-selection methods, 56.2\% of the problems
could be proved in~\cite{KaliszykU13b}. This was recently raised to
75.5\%~\cite{JakubuvCGKOP00U23}, mainly by using the
learning-guided E~\cite{Schulz13} (ENIGMA~\cite{JakubuvU17a,GoertzelCJOU21}) and Vampire~\cite{Vampire} (Deepire~\cite{DBLP:conf/cade/000121a}) systems.

Both E and Vampire are mainly saturation-style superposition systems. In the
recent years, instantiation-based systems and satisfiability modulo theories
(SMT) solvers such as cvc5~\cite{CVC5}, iProver~\cite{DBLP:conf/cade/Korovin08}
and Z3~\cite{z3} are however becoming competitive even for problems that do not
contain explicit theories in the SMT
sense~\cite{BlanchetteBP13,DesharnaisVBW22,GoertzelJKOPU22}. The problems that
they solve are often complementary to those solved by the superposition-based
systems.

\subsection{Contributions} 
In this work, we %
use %
instantiation-based methods (Section~\ref{sec:inst}) to solve automatically as many %
hard Mizar problems as possible.  Our main result is that the set of
ATP-provable MPTP problems has been increased by over \numprint{3000}, from 75.5\% to 80.7\%.
All these problems are proved by the cvc5 system which we improve in
several ways.  First, we use the Grackle system~\cite{HulaJJK22} (Section~\ref{sec:grackle}) to
automatically invent stronger strategies for MPTP (Section~\ref{sec:exp-grackle}). %
Our best strategy %
outperforms the previously best cvc5 strategy by 14\% and
our best 7-strategy portfolio solves 8.8\% more problems than the corresponding CASC portfolio (Section~\ref{sec:exp-time}).
We also combine strategy development with alternative
clausification methods. This turns out to have a surprisingly
high impact on the
instantiation-based system, contributing many new solutions (Section~\ref{sec:exp-cnf}).
Finally, we obtain further solutions by modifying the problems with premise selection (Section~\ref{sec:exp-premsel}).
Ultimately, these methods double the number of the previously ATP-unproved Mizar problems solved by cvc5 from
\numprint{1534} to \numprint{3021}. 
In this context, we consider a problem proved if there is at least one system that can solve it.
We show that the methods extend to previously unseen Mizar problems coming from newly added articled in a new version of MML (Section~\ref{sec:exp-transfer}).
We analyze the invented strategies (Section~\ref{sec:analysis}) and discuss several hard Mizar problems proved by them (Section~\ref{app:mizar-probs}).

\section{Instantiation-Based Methods}\label{sec:inst}
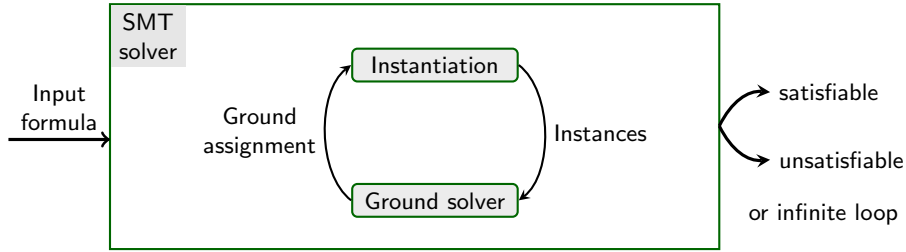
\begin{figure*}[t]
  \centering
  \usetikzlibrary{positioning}
  \begin{tikzpicture}[font=\sffamily,thick,xscale=1.35,yscale=1.8,%
    module/.style={draw=green!40!black,fill=gray!15,rounded corners=2pt}]
    \draw[green!40!black] (-2.2,-0.01) rectangle (3.8,1.8);
    \draw[very thick,->] (-3.2,.8) -- node[align=center,above] {Input\\formula} (-2.2,.8);

    \node[below right, align=center,fill=gray!20] at (-2.2,1.8) {SMT\\solver};
    \node[module,minimum width=2.2cm] (G) at (1,.35) {Ground solver};
    \node[module,minimum width=2.2cm] (I) at (1,1.35) {Instantiation};

    \draw[->,>=stealth] (I.east) to[bend left=60] node[right]{\footnotesize Instances}  (G.east);
    \draw[->,>=stealth] (G.west) to[bend left=60]
    node[align=center,left]{\footnotesize Ground\\ \footnotesize assignment} (I.west);

    \node[right] (s) at (4.3,1.15) {satisfiable};
    \node[right] (u) at (4.3,.65) {unsatisfiable};
    \node[right] at (4,.25) {or infinite loop};

    \draw[very thick,bend left,->,>=stealth] (3.8,0.9) to (s.west);
    \draw[very thick,bend right,->,>=stealth] (3.8,0.9) to (u.west);
  \end{tikzpicture}
  \caption{Schematic of an SMT solver with quantifier instantiation.}%
  \label{fig:schematic}
\end{figure*}

In contrast to saturation-style superposition systems, SMT solvers, namely cvc5, tackle quantifiers
by \emph{instantiations}, which can be seen as a direct application of the Herbrand's theorem.
A subformula
$(\forall x_1\dots x_n\,\phi)$ produces lemmas of the form $(\forall
x_1\dots x_n\,\phi)\rightarrow\phi[x_1/t_1,\dots,x_n/t_n]$,
with $\phi$ quantifier-free and $t_i$ ground terms. For example,
$\forall x\, R(f(x), c)$ may be instantiated as $(\forall x\, R(f(x), c))\rightarrow
R(f(c),c)$. Existential quantifiers are removed by Skolemization.

This approach consists of a loop alternating between a \emph{ground solver}
and an \emph{instantiation module} (Figure~\ref{fig:schematic}), where the
ground solver perceives quantified formulas as opaque propositions. After identifying a
model for the ground part, control shifts to the instantiation
module. This module generates new instances of the quantified sub-formulas that
are currently meant to hold, strengthening the grounded part of the formula.
The process stops if the ground part becomes unsatisfiable, if ever (model-based quantifier instantiation can also lead to satisfiable answers~\cite{ge-moura-cav09}).

The cvc5 solver implements several instantiation methods.
For decidable fragments, dedicated approaches exist, such as bit-vectors or linear arithmetic~\cite{DBLP:journals/fmsd/ReynoldsKK17,DBLP:journals/pacmpl/FarzanK18,DBLP:conf/cav/NiemetzPRBT18,DBLP:conf/lpar/BjornerJ15}.
Some of those can be seen as syntactic-driven approaches, \emph{e-matching}~\cite{DetlefsNS05,MoskalLK08} or syntax-guided instantiation~\cite{niemetz-tacas21}.
Other methods are semantic-driven such as \emph{model-based}~\cite{ge-moura-cav09} or \emph{conflict-based}~\cite{reynolds-fmcad14}.
A straightforward, but complete, approach for FOL is \textit{enumerative instantiation}~\cite{janota2021fair,ReynoldsBF18}, which exhaustively goes through all ground terms in the ground part.
Instantiation itself can also be guided by ML methods~\cite{janota-sat22}.

\section{Grackle: Targeted Strategy Invention for cvc5}
\label{sec:grackle}

Grackle~\cite{HulaJJK22} is a system for the automated invention of a portfolio of
solver strategies targeted to selected benchmark problems.
A user provides a set of benchmark problems and Grackle can automatically
discover a set of diverse solver strategies that maximize the number of solved
benchmark problems.
Grackle supports the invention of good-performing strategies for several solvers,
including ATP solvers E~\cite{Schulz13}, Vampire~\cite{Vampire},
Lash~\cite{BrownK22}, and an SMT solver Bitwuzla~\cite{Bitwuzla}.
Support for additional solvers can be easily added by providing a
parametrization of the solver strategy space, and by implementing a simple
wrapper to launch the solver.
In this paper, we extend Grackle to support an SMT solver
cvc5~\cite{CVC5}, and we evaluate its capabilities on a
first-order translation of Mizar problems.

Grackle is a successor of BliStr~\cite{blistr}, with which Grackle
shares the core of the strategy invention algorithm.
Grackle, however, generalizes the algorithm for an arbitrary solver.
BliStr/Grackle starts with user-provided solver strategies and 
interleaves a \emph{strategy evaluation} with a \emph{strategy invention} phase.
During the strategy evaluation phase, all available strategies are evaluated on
all benchmark problems, typically with some higher resource limit $T$.
This evaluation partitions the benchmark problems by individual strategy
performance, giving us, for each strategy $S$, the set of problems $P_S$ where
$S$ performs best.
The best strategy $S$ is then \emph{specialized} on problems $P_S$ in the
follow-up strategy invention phase in order to search for a strategy $S'$ with
an increased performance on~$P_S$.
This is achieved by launching an external parameter tuning software, like
ParamILS~\cite{ParamILS-JAIR} or SMAC3~\cite{lindauer2021smac3}, on problems
$P_S$ with the strategy $S$ as the initial starting point.
Moreover, a lower resource limit $t$ than in the evaluation phase ($T$) is
imposed on the solver during the tuning in order to guide the tuner
towards an improved performance on $P_S$.
The core idea, %
verified in previous
research~\cite{blistr,JakubuvU17,JakubuvSU17,HulaJJK22}, is that improved
performance on $P_S$ will bring about an improvement on other not-yet-solved
problems as well.
A new evaluation phase then proceeds with the extended portfolio.
Grackle has been extensively described~\cite{HulaJJK22} and we refer
the reader therein for a detailed exposition.

To use cvc5 with Grackle requires providing a parametrization of
the cvc5 strategy space.
A strategy for cvc5 is specified as command line options and their values.
While cvc5 supports more than \numprint{400} different options, we select all options
with non-numeric values relevant to problems in the theory of uninterpreted
functions (UF) with quantifiers.
This choice is guided by our indented application on the Mizar benchmark
problems which are expressed in the UF theory with a large number of quantified
formulae.
The cvc5 solver divides its options between \emph{regular} and \emph{expert}.
Hence we construct two parametrizations of cvc5 strategy space, one smaller with
the regular options only, and the second one with both regular and expert
options.
The regular parametrization has \numprint{98} parameters and the strategy space
covers about $10^{35}$ different strategies, while the full parametrization has
\numprint{168} parameters and the space size is about $10^{58}$.
As an exception, one of the expert options, namely
\verb|--cbqi-vo-exp|, was used also in the regular strategy space, to accommodate
all the options from the CASC strategies in both spaces.
We automatically extract all the options and their values from
cvc5's source files
\href{https://github.com/cvc5/cvc5/blob/cvc5-1.1.1/src/options/decision_options.toml}{decision\_options.toml},
\href{https://github.com/cvc5/cvc5/blob/cvc5-1.1.1/src/options/prop_options.toml}{prop\_options.toml},
\href{https://github.com/cvc5/cvc5/blob/cvc5-1.1.1/src/options/quantifiers_options.toml}{quantifiers\_options.toml},
\href{https://github.com/cvc5/cvc5/blob/cvc5-1.1.1/src/options/smt_options.toml}{smt\_options.toml}, and
\href{https://github.com/cvc5/cvc5/blob/cvc5-1.1.1/src/options/uf_options.toml}{uf\_options.toml}.%
\footnote{\url{https://github.com/cvc5/cvc5/tree/cvc5-1.1.1/src/options}}

Grackle additionally allows to express dependencies among options and thus to
describe options that are effective only under specific settings of another
option.  We automatically construct some dependencies from common prefixes of
option names, for example, the option \verb|--cbqi-mode| is applicable only
when the option \verb|--cbqi| is turned on.  While many of the dependencies
might be left unspecified, and while many of the options might be unrelated to
our benchmark problems, we leave this problem to Grackle and to the underlying
parameter tuner to deal with.  In this way, we also test Grackle's abilities to
deal with redundancies in the strategy space.
The cvc5 strategy space for Grackle can be found in the Grackle repository.%
\footnote{\url{https://github.com/ai4reason/grackle/tree/v0.2/grackle/trainer/cvc5}}

\section{Experiments}
\label{sec:exp}

\subsection{Dataset}
\label{sec:dataset}

Our goal is to prove as many of the remaining ATP-unproved
MPTP problems as possible. Of the \numprint{57880} problems, \numprint{43717} have
been proved\footnote{\url{https://github.com/ai4reason/ATP_Proofs}} in total in the previous experiments~\cite{JakubuvCGKOP00U23,KaliszykU13b}, thus,
\numprint{14163} problems remain to be proved.
Our strategy invention methods work by
gradually developing strategies that are faster and faster on 
solvable problems.
That is why we extend the set of the \numprint{14163}
ATP-unproved problems by another \numprint{4283} hard problems that were proved
only in the latest stages of the previous ATP experiments. We will use their versions with heuristically minimized
premises (using \emph{subproblem based
  minimization}~\cite{JakubuvCGKOP00U23}) to increase the chances of
the ATP systems. We also remove from this set \numprint{1585} problems for which the minimization was not done yet.\footnote{These are the non-theorem Mizar toplevel lemmas, for which the subproblem look-up (and thus also minimization) is more challenging.}
This
results in a set of \numprint{16861} hard problems on which we develop our
strategies.
These problems are by default in the FOF
format. We denote them \minFof  below. Later on (Section~\ref{sec:exp-cnf}), we apply different
clausifications to them. In Section~\ref{sec:exp-premsel}  we additionally
experiment with different premise selections for them.

\subsection{Overview of the Experiments}

Most of the experiments in this paper are conducted on the set of
\numprint{16861} hard Mizar problems described in Section~\ref{sec:dataset}.
Section~\ref{sec:exp-grackle} focuses solely on describing three Grackle runs
performed to develop a robust portfolio of cvc5 strategies specialized for
Mizar problems. All strategies are evaluated with a time limit of $30$ seconds.
Since increasing the time limit can still yield significant improvements,
selected strategies are evaluated in Section~\ref{sec:exp-time} with a higher
time limit, namely $600$ seconds. Sections~\ref{sec:exp-grackle} and 
\ref{sec:exp-time} use the same version of problems (\minFof) and differ
only in the time limit. In Section~\ref{sec:exp-cnf}, we explore different
clausification methods, and in Section~\ref{sec:exp-premsel}, we investigate
various premise selection methods. This implies that Sections~\ref{sec:exp-cnf}
and~\ref{sec:exp-premsel} use syntactically different but semantically
equivalent versions of Mizar problems. On the other hand,
Section~\ref{sec:exp-transfer} attempts to assess the overfitting of
Grackle-invented strategies on a new version of MML, thus the numbers reported
therein are not directly comparable with those in previous sections since the
problem sets differ.

\subsection{Experiments with Grackle Strategy Invention}
\label{sec:exp-grackle}

We evaluate%
\footnote{On two AMD EPYC 7513 32-Core processors @ 3680 MHz and with 514 GB RAM.}
the Grackle's ability to invent good-performing strategies for cvc5
on the \minFof  benchmark.
As a baseline, we consider all \numprint{16} strategies used in the cvc5's CASC
competition script (see Table~\ref{tab:casc-strats}).
We evaluate these \numprint{16} strategies with a \numprint{30}-second time
limit per strategy and problem.
The best strategy solves \numprint{2508} problems, while all the strategies together
solve  \numprint{3460} problems.
The two most complementary strategies are used as
the initial portfolio for the first Grackle run.

We perform consequently three Grackle runs, each with an overall timeout of 7
days.
Grackle terminates when all strategies have been already specialized, or when the
time is exhausted.
All three runs were terminated by timeout.
\begin{description}
   \item[(run \#1)]
      The first Grackle run starts with the two most complementary CASC
      strategies and uses the \emph{regular} strategy space (see
      Section~\ref{sec:grackle}).  
      The first run terminated after \numprint{7} days with \numprint{50} new strategies solving together
      \numprint{3459} problems with a $30$-second time limit per strategy and problem.
      Out of these problems, \numprint{345} are not solved by any of the $16$ baseline CASC
      strategies.
   \item[(run \#2)]
      The second run uses the same regular strategy space as run \#1 but it
      starts from the best $6$ strategies found in the first Grackle run.
      These initial strategies solve \numprint{3425} problems and Grackle invented \numprint{45}
      new strategies solving together \numprint{3696} problems
      with \numprint{485} unsolved by baseline strategies.
   \item[(run \#3)]
      The third run uses the same setup as run \#2 but it uses the full
      strategy space instead of the regular one.  Grackle invented \numprint{48} new
      strategies solving together \numprint{3856} problems with \numprint{629} 
      unsolved by baseline strategies.
\end{description}

\begin{figure*}[t]
   \begin{center}
      \includegraphics[width=1.0\columnwidth]{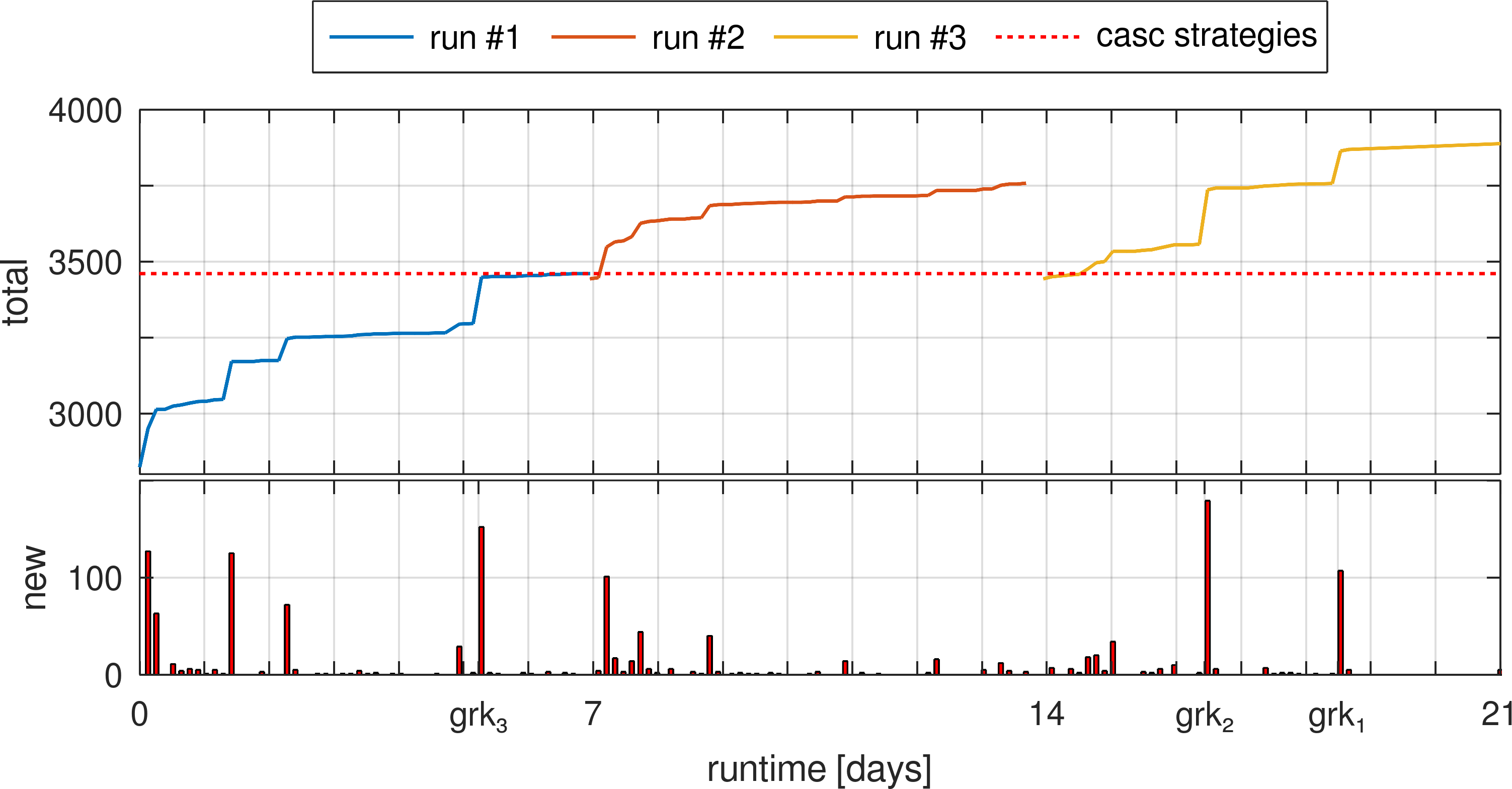}
\end{center}
\caption{Time progress of solved problems in Grackle runs.}%
\label{fig:result-grackle}
\end{figure*}

The time progress of the Grackle runs is visualized in
Figure~\ref{fig:result-grackle}.  The lower part (\emph{new}) shows the number
of new problems solved by the strategy invented at that time, while the upper
part (\emph{total}) shows the total number of problems solved.  The red dotted
line marks the performance of the $16$ baseline CASC strategies.  The $x$-axis
additionally shows the time of invention of the best three strategies
(\grk{i}).

The figure shows that the expert options added in run \#3 helped to develop stronger
strategies and to improve the results.  Strong strategies are sometimes
invented after several days of stagnation.  Grackle invented $143$ strategies,
which together solve \numprint{4113} problems.  The
best $16$ Grackle strategies solve \numprint{4039}, which increases the number of
\numprint{3460} problems solved by the $16$ baseline
strategies by $16.7\%$.  The best single strategy solves \numprint{2796}, which improves
on the best baseline strategy (which solves \numprint{2508}) by $11.5\%$. 

Additional data from the experiments are depicted in Table~\ref{tab:result-grackle}.
Columns \emph{solved} describe the initial and the final count of solved
problems together with the improvement over the initial strategies (\emph{new}), and
over the $16$ baseline CASC strategies (\emph{casc+}).
The column \emph{single best} states the number of problems solved by the best
invented strategy.
Columns \emph{strategies} describe the \emph{initial} count of strategies, the count of \emph{new} strategies, and how many strategies are needed to cover the \emph{final} number of solved problems (the length of the greedy cover sequence).
Columns \emph{specializations} provide the number of attempted specializations (\emph{total}) and \emph{failed} specializations where the output strategy was already known.

\begin{table}[t]
\begin{center}
\def\arraystretch{1.2}%
\setlength\tabcolsep{0.4em}%
\begin{tabular}{l||rrrr|r|rrr|rr}
   \multirow{2}{*}{\textit{run}} & 
   \multicolumn{4}{c|}{\textit{solved}} & 
   \multicolumn{1}{c|}{\textit{single}} &
   \multicolumn{3}{c|}{\textit{strategies}} & 
   \multicolumn{2}{c}{\textit{specializations}}
\\ 
   & 
   \textit{initial} & \textit{final} & \textit{new} & \textit{casc+} & 
   \multicolumn{1}{c|}{\textit{best}} &
   \textit{initial} & \textit{new} & \textit{needed} &
   \textit{total} & \textit{failed} 
\\\hline
\#1 &  2823 & 3459 & +636 & +345 & 2696 & 2 & 50 & 28 & 56 & 6  \\
\#2 &  3425 & 3696 & +271 & +482 & 2696 & 6 & 45 & 27 & 56 & 11  \\
\#3 &  3425 & 3856 & +431 & +629 & 2796 & 6 & 48 & 29 & 58 & 10
\end{tabular}
\end{center}
\caption{Grackle strategy invention for cvc5 on Mizar problems (Section~\ref{sec:exp-grackle}).}%
\label{tab:result-grackle}
\end{table}

\subsection{Experiments with Higher Time Limits}
\label{sec:exp-time}

The above (Section~\ref{sec:exp-grackle}) strategies were evaluated
in a \qty{30}{\second} time limit.  To solve more Mizar problems we proceed by
evaluating the best strategies with a higher time limit of \qty{600}{\second}.
Evaluation of a single strategy with this time limit takes about \qty{20}{hours}.
Hence, we evaluate strategies selectively as follows.  We start with the
strategies evaluated in \qty{30}{\second} and construct their greedy cover. %
The best \qty{30}{\second} strategy is then evaluated in \qty{600}{\second}.  The newly
evaluated strategy is then added to the greedy cover and the process is
iterated until new problems are being proved.
The best Grackle strategy \grk{1} solves \numprint{3496} problems, while the best CASC
strategy solves \numprint{3059}. This is a 14.3\% improvement.  We ended up with $7$ Grackle strategies and with $7$
CASC strategies evaluated in $600$s solving together \numprint{4653} problems (\numprint{4398}
by Grackles and \numprint{4043} by CASCs).  Together with the strategies evaluated in
the lower time limit we solved altogether \numprint{5035} of the benchmark problems at
this point.

This experiment with higher time limits also shows an interesting difference in
behavior between saturation-based ATP provers and instantiation-based SMT
solvers. While SMT solvers seem to benefit significantly from higher time
limits, ATP solvers typically benefit much less from it. For example, in the
comparable single-strategy setting, cvc5 solves almost $50\%$ more problems
when the time limit is increased from 60 to 600 seconds (see columns
\emph{alone} of \grk{1} in version \sBushy in
Table~\ref{tab:result-greedy-tptp} and
Table~\ref{tab:result-full-greedy-cover}). However, E Prover in the \emph{auto}
mode solves only $11\%$ more problems on the same benchmark problems with the
same increase of the time limit.
A similar relative performance gap is observed across different benchmarks.

\subsection{Experiments with Clausification Methods}
\label{sec:exp-cnf}
The Mizar problems are given as
TPTP~\cite{SutcliffeSY94} problems in first-order logic (FOF). For cvc5 we
translate them to the SMT2 language~\cite{barrett2010smt} in the theory of
uninterpreted functions (UF).  By default, cvc5 converts to
clausal normal form (CNF) internally but since
instantiation-based heuristics seem sensitive to problem
reformulation, we also experiment with external clausification.
This gives us syntactically different variants of the problems and
we can test whether cvc5 benefits from such alternative ways of
clausification.

We use E as the external clausifier and we construct
two more problem variants \minCnfA and \minCnfB. The \minCnfA version is
produced by using E's default clausification parameters, while for
\minCnfB we use a much more aggressive introduction of definitions for
frequent subformulas.
In particular, this is achieved by the E Prover's option
\texttt{--definitional-cnf} with values $24$ (default) and $4$, respectively.
The value indicates how many times a subformula needs to appear for a new
definition to be introduced via a new constant.
The \minCnfB methods
compared to \minCnfA almost halve the average number of literals in
the problems (\numprint{368.3} vs \numprint{668.2}) and the average number of symbols drops
to 60\% (\numprint{2512.7} vs \numprint{4124.1}).

The best Grackle strategy then solves \numprint{3231} problems in \qty{600}{\second} compared to \numprint{3125} by the best
CASC strategy.  Both these results use the \minCnfA clausification.  While
the individual performance of the strategies on the externally clausified problems is
lower than on the FOF variants, they are indeed highly complementary.
Eight Grackle strategies and six CASC strategies evaluated in
\qty{600}{\second} %
increase the number of the solved hard problems from \numprint{5035} to
\numprint{5404}.

\subsection{Experiments on Premise Selection Slices}
\label{sec:exp-premsel}

Based on the success with such problem reformulation, we perform additional
experiments, this time with different premise selection methods developed in our
prior work~\cite{JakubuvCGKOP00U23}.
Namely, we evaluate Grackle and baseline strategies on the \sBushy (i.e., not subproblem-minimized) variants of the
problems, on the strongest GNN (graph neural network~\cite{OlsakKU20}) premise selection slices with the threshold $-1$ (denoted here \sMirek), and
on LightGBM~\cite{LightGBM} premise selection slices with the threshold $0.1$ (\sPepa).
These variants were found complementary in
our previous experiments~\cite{JakubuvCGKOP00U23}. In a nutshell, the trained GNN puts at all the available premises into a large graph with the edges going between formulas, terms, subterms and symbols, runs several iterations of a \emph{message passing} algorithm in this large graph, and ultimately uses the aggregated information for deciding which of the premises are relevant for the conjecture. The LightGBM instead trains many decision trees on suitable features characterizing the premises to determine their relevance. These methods work quite differently and also sometimes recommend premises that are quite different from the ones used by human formalizers.

\begin{table}[t]
\small 
\begin{center}
\def\arraystretch{1.0}%
\setlength\tabcolsep{0.4em}%
\begin{tabular}{ll|lll|ll}
   \emph{version} & \emph{strat} & \multicolumn{2}{|c}{\emph{addon}} & \emph{total} & \emph{alone} & \emph{new}\\\hline
\minFof & \grk{1} & +3034 & - & 3034 & 3034       & 968 \\
\sMirek & \grk{1} & +521 & +17.17\% & 3555 & 1024 & 336      \\
\minFof & \grk{3} & +486 & +13.67\% & 4041 & 2828 & 887      \\
\sPepa & \grk{1} & +264 & +6.53\% & 4305 & 1260   & 405    \\
\minFof & \grk{2} & +187 & +4.34\% & 4492 & 2772  & 904     \\
\sBushy & \grk{2} & +177 & +3.94\% & 4669 & 963   & 329    \\
\minFof & \casc{10} & +73 & +1.56\% & 4742 & 2175 & 598      \\
\minFof & \casc{13} & +58 & +1.22\% & 4800 & 2348 & 666      \\
\sMirek & \grk{3} & +46 & +0.96\% & 4846 & 930    & 280   \\
\sPepa & \grk{2} & +30 & +0.62\% & 4876 & 1183    & 384   \\
\minFof & \casc{14} & +28 & +0.57\% & 4904 & 2650 & 828      \\
\sBushy & \casc{13} & +25 & +0.51\% & 4929 & 795  & 239     \\
\sPepa & \grk{3} & +24 & +0.49\% & 4953 & 1074    & 313   \\
\sMirek & \grk{2} & +23 & +0.46\% & 4976 & 1000   & 336    \\
\sBushy & \casc{14} & +17 & +0.34\% & 4993 & 913  & 302     \\
\sBushy & \casc{10} & +11 & +0.22\% & 5004 & 609  & 188     \\
\sBushy & \grk{1} & +9 & +0.18\% & 5013 & 962     & 319  \\
\sBushy & \grk{3} & +8 & +0.16\% & 5021 & 746     & 195  \\
\sMirek & \casc{13} & +7 & +0.14\% & 5028 & 899   & 287    \\
\sPepa & \casc{10} & +7 & +0.14\% & 5035 & 895    & 262   \\
\sMirek & \casc{14} & +6 & +0.12\% & 5041 & 954   & 313    \\
\sPepa & \casc{13} & +5 & +0.10\% & 5046 & 1051   & 309    \\
\sPepa & \casc{14} & +4 & +0.08\% & 5050 & 1137   & 363    \\
\end{tabular}
\end{center}
\caption{Full greedy cover on FOF slices \minFof, \sBushy, \sMirek, and \sPepa with $60$s timeout.}%
\label{tab:result-greedy-tptp}
\normalsize
\end{table}

We again evaluate our strategies on such problems, %
first with a lower (\qty{60}{\second}) time limit.
Table~\ref{tab:result-greedy-tptp} shows a comparative evaluation of the \qty{60}{\second} Grackle and
CASC strategies on these slices.
In the table, the column \emph{version} displays the benchmark version and
the column \emph{strat} is the strategy name.
The column \emph{addon} describes the addition of the strategy to the portfolio,
that is, it lists the number of problems added, and the same in percents.
The column \emph{total} lists the cumulative performance of the portfolio up to
that line.
Finally, the column \emph{alone} shows the individual performance of the
strategy and the column \emph{new} shows the number of Mizar problems unproved
in our previous
research~\cite{JakubuvCGKOP00U23,KaliszykU13b}.
The specification of Grackle-invented strategies (\grk{i}) can be found in
Table~\ref{tab:best-five} while the definitions of CASC strategies is in
Table~\ref{tab:casc-strats}.
These strategies are further analyzed in Section~\ref{sec:analysis}.

Based on this, we evaluate %
the best Grackle strategy \grk{1} on all three slices
in \qty{600}{\second}.  This alone raises the number of solved
problems from \numprint{5404} to \numprint{6363}. After adding also
the \qty{60}{\second} results, we obtain in total \numprint{6469} hard
problems solved, of which \textbf{\numprint{3021}} were not proved by
ATPs before.\footnote{\label{note:repo}
   The lists of problems solved by the individual strategies and the strategy definitions
are available at \url{https://github.com/ai4reason/cvc5_grackle_mizar}.} This
is our main result. We have proved \textbf{21.3\%}
of the remaining ATP-unproved problems, and increased the total number
of \emph{all} ATP-proved Mizar problems to \numprint{46738}
(\textbf{\qty{80.7}{\percent}}).  About half of the \numprint{3021}
problems (\numprint{1534}) can be solved by the cvc5 CASC
strategies. For the remaining half, some of our methods (new
strategies, different clausifications or premise slices) are necessary.

The first $10$ strategies in the final greedy
cover are shown in Table~\ref{tab:result-greedy-cover} (left).
The meaning of the columns is the same as in Table~\ref{tab:result-greedy-tptp}, that is, the column \emph{version} displays benchmark version,
the column \emph{strat} is the strategy name,
the column \emph{addon} describes the addition of the strategy to the portfolio,
and \emph{alone} shows the individual performance of the
strategy.
We can see that the Grackle-invented strategies clearly dominate the greedy
cover.
While premise selection slices \sMirek, \sPepa, and \sBushy exhibit low
individual performance, they provide many new solutions.
This is often due to the alternative proofs proposed by the premise selection methods trained over many previous proofs.

For the sake of completeness, Table~\ref{tab:result-full-greedy-cover} additionally presents extended results.
The table mixes strategies evaluated with different time limits denoted in the
column \emph{timeout}.
The meaning of other columns is the same as in Table~\ref{tab:result-greedy-tptp}.
All Grackle and CASC strategies solve together \numprint{6363} Mizar problems.%
\footnote{The strategies not listed in Table~\ref{tab:best-five} (like \grk{\mathtt{169baa}})
   can be found in our repository (Note~\ref{note:repo}).
}

\begin{table}[t]
\footnotesize
\begin{center}
\def\arraystretch{1.1}%
\setlength\tabcolsep{0.8em}%
\begin{tabular}{ll|ll||ll|ll}
\multicolumn{4}{c||}{Results on MML} &
\multicolumn{4}{c}{Transfer to new MML} 
\\\hline
\emph{version} & \emph{strat} & \emph{addon} & \emph{alone} &
\emph{version} & \emph{strat} & \emph{addon} & \emph{alone}
\\\hline
\minFof     & \grk{1}   & +3496 & 3496 &  \textit{cnf1} & \grk{2}   &  +4861 & 4861 \\
\minCnfA    & \grk{2}   & +738  & 3231 &  \textit{fof}  & \grk{1}   &  +433  & 4541 \\
\sMirek     & \grk{1}   & +535  & 1215 &  \textit{cnf1} & \grk{3}   &  +164  & 4495 \\
\sBushy     & \grk{1}   & +311  & 1441 &  \textit{fof}  & \casc{13} &  +78   & 4406 \\
\minFof     & \grk{3}   & +298  & 3220 &  \textit{fof}  & \grk{3}   &  +53   & 4195 \\
\sPepa      & \grk{1}   & +233  & 1512 &  \textit{fof}  & \grk{2}   &  +39   & 4418 \\
\minCnfA    & \grk{3}   & +161  & 3223 &  \textit{cnf1} & \grk{1}   &  +33   & 4811 \\
\minCnfA    & \casc{10} & +112  & 3125 &  \textit{cnf1} & \casc{10} &  +17   & 4211 \\
\minFof     & \grk{2}   & +90   & 3146 &  \textit{cnf2} & \grk{1}   &  +14   & 4417 \\
\minCnfB    & \grk{2}   & +62   & 2949 &  \textit{fof}  & \casc{10} &  +12   & 3952 \\
\end{tabular}
\end{center}

\caption{Results on MML (left),
   transfer to new MML (right).}%
\label{tab:result-greedy-cover}
\normalsize
\end{table}

\begin{table}[p]
\small 
\begin{center}
\def\arraystretch{1.0}%
\setlength\tabcolsep{0.4em}%
\begin{tabular}{lll|lll|ll}
   \emph{version} & \emph{timeout} & \emph{strat} & \multicolumn{2}{|c}{\emph{addon}} & \emph{total} & \emph{alone} & \emph{new}\\\hline
\minFof & 600 & \grk{1} & +3496 & - & 3496 & 3496  &                    1243  \\
\minCnfA & 600 & \grk{2} & +738 & +21.11\% & 4234 & 3231  &             1192         \\
\sMirek & 600 & \grk{1} & +535 & +12.64\% & 4769 & 1215  &              432         \\
\sBushy & 600 & \grk{1} & +311 & +6.52\% & 5080 & 1441  &               553        \\
\minFof & 600 & \grk{3} & +298 & +5.87\% & 5378 & 3220  &               1146       \\
\sPepa & 600 & \grk{1} & +233 & +4.33\% & 5611 & 1512  &                541       \\
\minCnfA & 600 & \grk{3} & +161 & +2.87\% & 5772 & 3223  &              1092        \\
\minCnfA & 600 & \casc{10} & +112 & +1.94\% & 5884 & 3125  &            999           \\
\minFof & 600 & \grk{2} & +90 & +1.53\% & 5974 & 3146  &                1131      \\
\minCnfB & 600 & \grk{2} & +62 & +1.04\% & 6036 & 2949  &               1045       \\
\minFof & 600 & \grk{5} & +49 & +0.81\% & 6085 & 3086  &                1063      \\
\minCnfA & 600 & \grk{1} & +35 & +0.58\% & 6120 & 3163  &               1110       \\
\minCnfB & 600 & \grk{5} & +31 & +0.51\% & 6151 & 2909  &               1030       \\
\minCnfA & 600 & \grk{5} & +27 & +0.44\% & 6178 & 3113  &               1099       \\
\minCnfB & 600 & \grk{3} & +22 & +0.36\% & 6200 & 2851  &               934        \\
\minFof & 600 & \casc{13} & +16 & +0.26\% & 6216 & 2711  &              848         \\
\minCnfB & 600 & \casc{10} & +14 & +0.23\% & 6230 & 2695  &             787          \\
\minFof & 600 & \casc{10} & +13 & +0.21\% & 6243 & 2575  &              795         \\
\minFof & 600 & \grk{\mathtt{169baa}} & +12 & +0.19\% & 6255 & 2993  &  722                     \\
\minCnfA & 600 & \casc{06} & +12 & +0.19\% & 6267 & 2334  &             1002         \\
\minFof & 600 & \casc{09} & +11 & +0.18\% & 6278 & 1064  &              150         \\
\minFof & 600 & \casc{14} & +10 & +0.16\% & 6288 & 3059  &              1057        \\
\minFof & 30 & \grk{\mathtt{d73c5e}} & +9 & +0.14\% & 6297 & 2901  &    986                   \\
\minFof & 600 & \casc{06} & +8 & +0.13\% & 6305 & 2380  &               716        \\
\minFof & 30 & \grk{\mathtt{393769}} & +7 & +0.11\% & 6312 & 2671  &    803                   \\
\minCnfA & 600 & \casc{13} & +7 & +0.11\% & 6319 & 2948  &              977         \\
\minFof & 600 & \casc{07} & +5 & +0.08\% & 6324 & 2955  &               916        \\
\minFof & 600 & \casc{16} & +5 & +0.08\% & 6329 & 2976  &               885        \\
\minFof & 30 & \grk{\mathtt{1fe2d9}} & +5 & +0.08\% & 6334 & 2770  &    992                   \\
\minCnfB & 600 & \casc{13} & +5 & +0.08\% & 6339 & 2726  &              968         \\
\minFof & 30 & \grk{\mathtt{014565}} & +3 & +0.05\% & 6342 & 2666  &    849                   \\
\minFof & 30 & \grk{\mathtt{043c34}} & +3 & +0.05\% & 6345 & 2544  &    769                   \\
\minCnfB & 600 & \casc{06} & +3 & +0.05\% & 6348 & 2090  &              631         \\
\minCnfB & 600 & \grk{1} & +3 & +0.05\% & 6351 & 2817  &                933       \\
\minFof & 600 & \grk{4} & +2 & +0.03\% & 6353 & 3320  &                 859      \\
\minFof & 30 & \grk{\mathtt{166bee}} & +2 & +0.03\% & 6355 & 2671  &    1163                  \\
\minFof & 30 & \grk{\mathtt{04f79f}} & +1 & +0.02\% & 6356 & 2484  &    725                   \\
\minFof & 30 & \grk{\mathtt{0f4750}} & +1 & +0.02\% & 6357 & 2465  &    723                   \\
\minFof & 30 & \grk{\mathtt{1499bd}} & +1 & +0.02\% & 6358 & 2238  &    641                   \\
\minFof & 30 & \grk{\mathtt{1afb4a}} & +1 & +0.02\% & 6359 & 463  &     41                   \\
\minFof & 30 & \grk{\mathtt{340075}} & +1 & +0.02\% & 6360 & 2556  &    742                   \\
\minFof & 30 & \grk{\mathtt{52ae2f}} & +1 & +0.02\% & 6361 & 2670  &    800                   \\
\minFof & 30 & \grk{\mathtt{7dac18}} & +1 & +0.02\% & 6362 & 173  &     6                    \\
\minFof & 30 & \grk{\mathtt{ba0f42}} & +1 & +0.02\% & 6363 & 1810  &    509                   \\
\end{tabular}
\end{center}
\caption{Full complete greedy cover on MML problems.}%
\label{tab:result-full-greedy-cover}
\normalsize
\end{table}

\subsection{Transfer to New MML}
\label{sec:exp-transfer}
To assess the overfitting of the methods we evaluate the best
three Grackle and the best three CASC strategies on one more
benchmark.  We use \numprint{13370} \sBushy problems coming from newly
added articles in MML version \num{1382}. 
Table~\ref{tab:result-greedy-cover} (right) shows the results.
The Grackle strategies outperform all CASC strategies, even
though the improvement is smaller 
than on the MML problems they were developed for. Alternative
clausification methods again provide a considerable improvement.

\section{Analysis of the Invented Strategies}
\label{sec:analysis}
As usual with automated strategy invention, there are many new
combinations of parameters that may require deeper analysis to
understand the automatically invented behavior. That is why we make
them publicly
available.\footnote{See Note~\ref{note:repo}.}
As a baseline and as a starting point for Grackle strategy inventions, we
consider \numprint{16} strategies used in the cvc5's CASC competition script.%
\footnote{\url{https://github.com/cvc5/cvc5/blob/cvc5-1.1.1/contrib/competitions/casc/run-script-cascj11-fof}}
The strategies are listed in Table~\ref{tab:casc-strats}.
The best Grackle strategies are depicted in Table~\ref{tab:best-five}.

\begin{table}[t]
\small 
\begin{center}
\def\arraystretch{1.2}%
\setlength\tabcolsep{0.4em}%
\begin{tabular}{l|l}
   name & cvc5 strategy options
\\\hline\hline
\casc{1} & \verb|--decision=internal --simplification=none --no-inst-no-entail| \\
         & \verb|--no-cbqi --full-saturate-quant| \\\hline
\casc{2} & \verb|--no-e-matching --full-saturate-quant| \\\hline
\casc{3} & \verb|--no-e-matching --enum-inst-sum --full-saturate-quant| \\\hline
\casc{4} & \verb|--finite-model-find --uf-ss=no-minimal| \\\hline
\casc{5} & \verb|--multi-trigger-when-single --full-saturate-quant| \\\hline
\casc{6} & \verb|--trigger-sel=max --full-saturate-quant| \\\hline
\casc{7} & \verb|--multi-trigger-when-single --multi-trigger-priority| \\
         & \verb|--full-saturate-quant| \\\hline
\casc{8} & \verb|--multi-trigger-cache --full-saturate-quant| \\\hline
\casc{9} & \verb|--prenex-quant=none --full-saturate-quant| \\\hline
\casc{10} & \verb|--enum-inst-interleave --decision=internal --full-saturate-quant| \\\hline
\casc{11} & \verb|--relevant-triggers --full-saturate-quant| \\\hline
\casc{12} & \verb|--finite-model-find --e-matching --sort-inference --uf-ss-fair| \\\hline
\casc{13} & \verb|--pre-skolem-quant=on --full-saturate-quant| \\\hline
\casc{14} & \verb|--cbqi-vo-exp --full-saturate-quant| \\\hline
\casc{15} & \verb|--no-cbqi --full-saturate-quant| \\\hline
\casc{16} & \verb|--macros-quant --macros-quant-mode=all --full-saturate-quant| \\\hline
\end{tabular}
\end{center}
\caption{CASC baseline strategies used in the experiments.}%
\label{tab:casc-strats}
\normalsize
\end{table}

\begin{table}[ht]
\small 
\begin{center}
\def\arraystretch{1.2}%
\setlength\tabcolsep{0.4em}%
\begin{tabular}{l|l}
   name & cvc5 strategy options
\\\hline\hline
\grk{1} & \verb|--cbqi-vo-exp --cond-var-split-quant=agg --full-saturate-quant | \\
        & \verb|--relational-triggers | \\\hline
\grk{2} & \verb|--cbqi-vo-exp --full-saturate-quant --miniscope-quant=off| \\
        & \verb|--multi-trigger-priority --no-static-learning --relevant-triggers| \\
        & \verb|--ieval=off| \\\hline
\grk{3} & \verb|--full-saturate-quant --multi-trigger-priority| \\
        & \verb|--multi-trigger-when-single --term-db-mode=relevant | \\\hline
\grk{4} & \verb|--cbqi-vo-exp --cond-var-split-quant=agg | \\
        & \verb|--full-saturate-quant --inst-when=last-call | \\\hline
\grk{5} & \verb|--cbqi-all-conflict --full-saturate-quant --inst-when=full-delay| \\
        & \verb|--macros-quant --multi-trigger-priority --quant-dsplit=none | \\
        & \verb|--quant-dsplit=none --trigger-sel=min-s-all --uf-ss=none | \\\hline
\end{tabular}
\end{center}
\caption{Best five strategies invented by Grackle.}%
\label{tab:best-five}
\normalsize
\end{table}

Interestingly, the different Grackle-invented strategies focus mainly on
changing the behavior of the different components of the quantifier
instantiation module of cvc5, cf.~Section~\ref{sec:inst}.
By default cvc5 relies on e-matching~\cite{DetlefsNS05,MoskalLK08}, which is incomplete, which also means
that the solver may quickly give up (return the output \texttt{unknown}). The option
\small\verb|--full-saturate-quant|\normalsize, runs the default mode but if that fails to answer,
the solver resorts to the enumerative mode (complete for FOL~\cite{ReynoldsBF18}). This explains why
this option is so prevalent in the invented strategies.

In \grk{1} and \grk{3}, e-matching's behavior is changed by changing the
trigger-generation policy. In \grk{1} and  \grk{2}, the option
\small\verb|--cbqi-vo-exp|\normalsize\xspace affects the behavior of the conflict-driven
instantiation~\cite{reynolds-fmcad14}.
The option \small\verb|--cond-var-split-quant|\normalsize\xspace affects the quantifier splitting
policy. The option \small\verb|--term-db-mode=relevant|\normalsize\xspace enforces a stricter policy on
ground term  filtering.
In general, it seems that the essence of a successful strategy is a combination
of enumerative instantiations with an appropriate trigger selection for
e-matching.
In the next section (Section~\ref{app:mizar-probs}) we discuss the influence of such options on the solution of several hard Mizar problems.

\section{Interesting Mizar Problems Proved}
\label{app:mizar-probs}

Since we are focusing on the 25\% of the Mizar problems that have not
been proved by ATPs so far, the newly solved problems are typically
quite involved, with long proofs both in Mizar and in cvc5. 127 of
them take more than 100 lines to prove in Mizar, and the average Mizar
proof length is 41. This is one page of a proof in a paper like this.

A previously ATP-unproved problem that seems relatively easy for many
of the cvc5 strategies is
\texttt{KURATO\_1:6}\footnote{\url{http://grid01.ciirc.cvut.cz/~mptp/7.13.01_4.181.1147/html/kurato_1.html\#T6}}
related to the well-known Kuratowski's closure-complement
problem.\footnote{\url{https://en.wikipedia.org/wiki/Kuratowski\%27s_closure-complement_problem}}
The theorem shows that for any set $A$, its \texttt{Kurat14Set} (i.e.,
a family of 14 sets created by applying closure and complement
operations in a particular way to $A$) is already closed under
complement and closure:
\begin{footnotesize}
\begin{verbatim}
definition
let T be non empty TopSpace;
let A be Subset of T;
func Kurat14Set A -> Subset-Family of T equals
{ A, A-, A-`, A-`-, A-`-`, A-`-`-, A-`-`-` } \/ 
{ A`, A`-, A`-`, A`-`-, A`-`-`, A`-`-`-, A`-`-`-` };
end;

theorem Th6: for T being non empty TopSpace
for A, Q being Subset of T st Q in Kurat14Set A holds
Q` in Kurat14Set A & Q- in Kurat14Set A; 
\end{verbatim}
\end{footnotesize}
The proof has 131 lines in Mizar, however it indeed seems 
achievable by instantiation-based methods that gradually enumerate the
applications of closure and complement to the skolems and use
congruence closure when a more complex term can be shown to be equal
to a less complex term. The problem is a combination of equational
reasoning and a large case split (14 cases), which is what 
likely makes it hard for the superposition-based systems. The success may
indicate that a full ATP (or AI/TP) solution of the Kuratowski's
closure-complement problem may not be too far today, because proposing
the \texttt{Kurat14Set} and finding automatically a suitable family of
14 distinct sets (to show that 14 is indeed the smallest number) also seems
within the reach of today's systems.

The problem
\texttt{ASYMPT\_1:18}\footnote{\url{http://grid01.ciirc.cvut.cz/~mptp/7.13.01_4.181.1147/html/asympt_1.html\#T18}}
is on the other hand only provable with a single Grackle-invented
strategy \grk{2} and external clausification, taking 62~s. The problem states that
the functions $f(n)= n \bmod 2$ and $g(n)=n+1 \bmod 2$ are not in
the Big O relation (in any direction).
\begin{footnotesize}
\begin{verbatim}
theorem
 for f,g being Real_Sequence st 
   (for n holds f.n = n mod 2) & (for n holds g.n = n+1 mod 2) 
  holds ex s,s1 being eventually-nonnegative Real_Sequence
  st s = f & s1 = g & not s in Big_Oh(s1) & not s1 in Big_Oh(s)
\end{verbatim}
\end{footnotesize}
The Mizar proof has 122 lines and again goes through several case
splits related to the $\bmod$
$2$ values. However a lot of knowledge (often equational) about the
arithmetical expressions, modulo and inequality has to be applied
too.\footnote{Note that the Mizar/MPTP translation translates
  everything as uninterpreted functions, i.e., there is no reliance on
  the arithmetical theories implemented in cvc5.} The fact that this
can be done by an instantiation-based system is quite remarkable, and
probably also due to the fact that the terms that arise in the proof
are not extremely complicated thanks to the $\{0,1\}$ codomain of the
functions involved. The option \texttt{--multi-trigger-priority} seems
indispensable for solving the problem, showing the importance of the
heuristics for handling instantiation triggers. This may be an
opportunity for further AI/ML methods learning even finer control of
the triggers in such systems.

Finally, theorem \texttt{ROBBINS4:3}\footnote{\url{http://grid01.ciirc.cvut.cz/~mptp/7.13.01_4.181.1147/html/robbins4.html\#T3}} shows an equivalent condition for ortholattices:
\begin{footnotesize}
\begin{verbatim}
for L being non empty OrthoLattStr holds L is Ortholattice iff 
  (for a, b, c being Element of L holds  
     (a "\/" b) "\/" c = (c` "/\" b`)` "\/" a)
& (for a, b being Element of L holds a = a "/\" (a "\/" b)) 
& for a, b being Element of L holds a = a "\/" (b "/\" b`)
\end{verbatim}
\end{footnotesize}
The problem can only be solved by the Grackle-invented strategy 89fc24
and it takes 137 s. The Mizar proof has 145 lines and uses a lot of
equational reasoning in lattice theory. It is quite surprising that a
proof with so much equality could not be done by the superposition
based systems, and that it can be done by cvc5. Again, triggers seem important here, together with the 
\texttt{--term-db-mode=relevant} option which further limits the sets of possible quantifier instantiations.

\section{Conclusions and Future Work}

We have solved \textbf{\numprint{3021}} (21.3\%) of the remaining
\numprint{14163} hard Mizar problems, raising the percentage of automatically
proved Mizar problems from 75.5\% to \textbf{80.7\%}. This was mainly done by
automatically inventing suitable instantiation-based strategies for the cvc5
solver, using our Grackle system. Further improvements were obtained by using
alternative clausifications of the problems, and also alternative premise
selections. Such problem transformations have a surprisingly large effect on
the instantiation-based procedures and are likely to be explored further when
creating strong portfolios for such systems.

The invented cvc5 strategies perform well also on a set of new
problems added in a later version of the Mizar library, showing only
limited overfitting. Given today's cvc5's good performance on corpora
such as Isabelle/Sledgehammer and TPTP, it may be also interesting to
repeat our strategy invention experiments for the TPTP problems and
for problems exported from various non-Mizar hammer systems. In
general, training instantiation-based systems in various ways is an
emerging research topic that may bring interesting improvements to
some of today's strongest ATP/SMT methods.

\section*{Acknowledgments}
Supported 
by the Czech MEYS under the ERC CZ project no.~LL1902 \emph{POSTMAN},
by Amazon Research Awards, 
by EU ICT-48 2020 project no.~952215 \emph{TAILOR}, 
by ERC PoC grant no.~101156734 \emph{FormalWeb3},
by CISCO grant no. 2023-322029, and
co-funded by the European Union under the project \emph{ROBOPROX}
(reg.~no.~CZ.02.01.01/00/22\_008/0004590).

\bibliographystyle{plain}

\bibliography{ate11,quant}
\newpage

\appendix

\end{document}